\documentclass[lettersize,journal]{IEEEtran}
\usepackage{float}
\usepackage{amsmath,amsfonts}
\usepackage[ruled,linesnumbered]{algorithm2e}
\usepackage{booktabs}
\usepackage{multirow}
\usepackage{makecell}
\usepackage{array}
\usepackage[caption=false,font=normalsize,labelfont=sf,textfont=sf]{subfig}
\usepackage{textcomp}
\usepackage{stfloats}
\usepackage{url}
\usepackage{verbatim}
\usepackage{graphicx}
\usepackage{cite}
\usepackage{xcolor}         
\usepackage[square,sort,comma,numbers]{natbib}

\begin{document}

\title{Lifelong Evolution: Collaborative Learning between Large and Small Language Models for Continuous Emergent Fake News Detection}

\author{Ziyi Zhou, Xiaoming Zhang,  Litian Zhang, Yibo Zhang\\ Zhenyu Guan,  Chaozhuo Li, and Philip S. Yu, \textit{Fellow, IEEE}
\thanks{

Ziyi Zhou, Xiaoming Zhang, Litian Zhang, Yibo Zhang and Zhenyu Guan are with School of Cyber Science and Technology, Beihang University, Beijing 100191, P. R. China (e-mail: ziyizhou@buaa.edu.cn; yolixs@buaa.edu.cn; litianzhang@buaa.edu.cn; 21371167@buaa.edu.cn; guanzhenyu@buaa.edu.cn). (Corresponding author: Xiaoming Zhang.)

Chaozhuo Li is with School of Cyber Science and Technology, Beijing University of Posts and Telecommunications,
Beijing 100876, China (e-mail:  lichaozhuo@bupt.edu.cn).

Philip S. Yu is with the Department of Computer Science, University of Illinois at Chicago, Chicago, IL 60607 USA (e-mail: psyu@uic.edu). ORCID: 0000-0002-3491-5968.
}}

\markboth{Journal of \LaTeX\ Class Files,~Vol.~14, No.~8, August~2021}%
{Shell \MakeLowercase{\textit{et al.}}: A Sample Article Using IEEEtran.cls for IEEE Journals}


\maketitle

\begin{abstract}
The widespread dissemination of fake news on social media has significantly impacted society, resulting in serious consequences. 
Conventional deep learning methodologies employing small language models (SLMs) suffer from extensive supervised training requirements and difficulties adapting to evolving news environments due to data scarcity and distribution shifts.
Large language models (LLMs), despite robust zero-shot capabilities, fall short in accurately detecting fake news owing to outdated knowledge and the absence of suitable demonstrations. 
In this paper, we propose a novel Continuous Collaborative Emergent Fake News Detection (C$^2$EFND) framework to address these challenges.
The C$^2$EFND framework strategically leverages both LLMs' generalization power and SLMs' classification expertise via a multi-round collaborative learning framework.  
We further introduce a lifelong knowledge editing module based on a Mixture-of-Experts architecture to incrementally update LLMs and a replay-based continue learning method to ensure SLMs retain prior knowledge without retraining entirely. 
Extensive experiments on Pheme and Twitter16 datasets demonstrate that C$^2$EFND significantly outperforms existed methods, effectively improving detection accuracy and adaptability in continuous emergent fake news scenarios.
\end{abstract}

\begin{IEEEkeywords}
Fake News Detection, Large Language Models, Continue Learning
\end{IEEEkeywords}

\section{Introduction}
\begin{figure*}[t]
\centering
\includegraphics[scale=1.0]{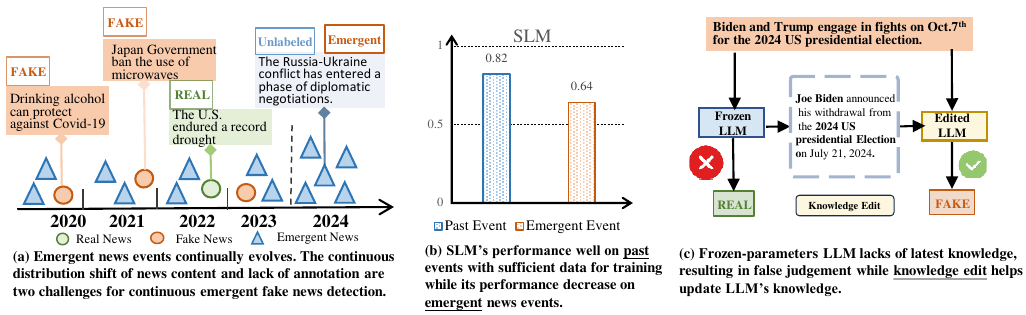}
    \setlength{\abovecaptionskip}{-1mm}
    \caption{Challenges in continuous emergent fake news detection: distribution shift, sparse labels, and static models.}
    \label{fig:intro_2}
    \vspace{-4mm}
\end{figure*}

\IEEEPARstart{T}{he} rampant spread of fake news on the Internet has already caused significant societal impact~\cite{olan2024fake}. For instance, the spread of fake news during the Covid-19 pandemic has led to harmful consequences such as drug misuse and incorrect treatment methods~\cite{van2020inoculating}. As illustrated in Figure~\ref{fig:intro_2}(a), fake news on emergent events evolves continuously, presenting a challenge for real-time detection systems to keep pace with its evolution. Furthermore, an alarming pattern known as ``rumor resurgence'' frequently occurs in social media, wherein past misinformation reappears, perpetuating its societal impact~\cite{shin2018diffusion}. This dynamic nature of fake news highlights the need for detection models that can continuously adapt to new events while retaining the ability to detect past events.

Traditional fake news detection approaches predominantly utilize small language models (SLMs) like BERT~\cite{devlin2018bert} to extract semantic features from news and train classifiers for fake news detection~\cite{szczepanski2021new}. To enable more accurate judgment by the models, some methods attempt to leverage external knowledge and propagation paths in social media to assist in detection and have shown promising results~\cite{zhang2024reinforced,jing2025dpsg}. However, these models are constrained by limited generalization and adaptability to new events due to the need for extensive re-training on sufficient annotated data, which is both time-consuming and labor-intensive~\cite{silva2021embracing}. Consequently, as Figure~\ref{fig:intro_2}(b) illustrates, the performance of SLMs significantly decreases when confronted with emergent news events, where labeled data is scarce and exhibits distribution shifts.

\begin{figure}[t]
\centering
\includegraphics[scale=1.0]{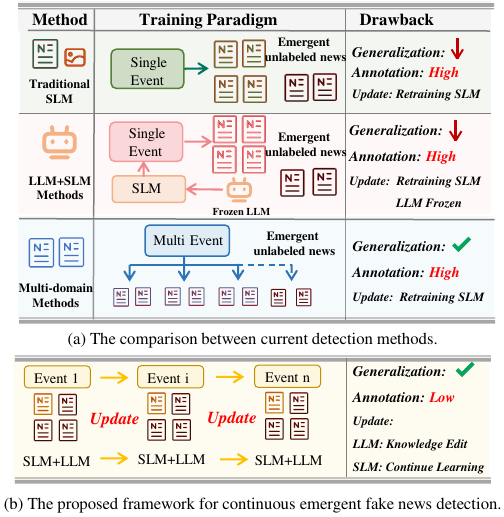}
    \setlength{\abovecaptionskip}{-1mm}
    \caption{(a) denotes the drawbacks of previous methods while (b) presents our proposed method.}
    \label{fig:intro_1}
    \vspace{-5mm}
\end{figure}
Large language models (LLMs) have demonstrated remarkable capabilities across various NLP tasks, demonstrating robust capabilities in language understanding and generation~\cite{chang2024survey,zhao2023survey}. However, previous studies have shown that LLMs perform poorly on fake news detection tasks~\cite{hu2024bad}. 
This might be attributed to two main reasons: Firstly, most LLMs used in fake news detection are frozen-parameter models, which lack the task-specific training required to accurately understand the nuances of fake news and identify deceptive content. Secondly, frozen-parameters LLMs do not receive continuous updates with the latest accurate knowledge, limiting their ability to detect emergent news events. 
Due to the limited detection performance of LLMs, recent researches mainly focus on leveraging LLMs as agents to help 
SLMs by providing additional knowledge~\cite{wan2024dell,wu2024fake}. For instance, Dell~\cite{wan2024dell} utilizes LLMs as agents to form a social network by generating user comments while conducting sentiment analysis to help SLMs in fake news detection. ARG~\cite{hu2024bad} leverages LLMs to analyze the content and provide  rationales for SLMs. Although these methods leveraging LLMs have effectively improved detection accuracy, they overlook two significant factors. On the one hand, they utilize LLMs merely to assist SLMs, ignoring the learning and potential detection capabilities of the LLMs. On the other hand, previous approaches utilize frozen-parameter LLMs, which prevents the models from understanding the latest news events, resulting in hallucination on emergent events as illustrated in Figure~\ref{fig:intro_2}(c). Furthermore, these methods still rely on a substantial amount of data to train SLMs and only focus on single-event, lacking the updating mechanism for both LLMs and SLMs as shown in Figure~\ref {fig:intro_1}(a).

Considering the diversity of news types, an increasing number of studies have shifted towards multi-domain fake news detection~\cite{nan2021mdfend,zhu2022memory}. 
These multi-domain methods aim to handle the diverse nature of fake news by capturing domain-specific features and adapting to varied content. However, as illustrated in Figure~\ref{fig:intro_1}(a), they still face two challenges in real-world applications. First, they still rely on abundant labeled data from multi-events for supervised learning. Second, while these methods attempt to address the multi-domain nature of fake news, they lack efficient mechanisms to update model knowledge in response to emergent events. Consequently, as shown in Figure~\ref{fig:intro_1}(a), they require frequent retraining on extensive datasets to maintain performance across past and emergent events, which is not feasible in fast-paced, dynamic environments such as social media~\cite{zhou2020survey}.
Therefore, there is an urgent need for a model with continue learning capability, able to adapt to an evolving news environment using only a small amount of data while maintaining accurate detection abilities on past events without full retraining.

In this paper, we propose a novel collaborative learning framework between LLMs and SLMs for continuous emergent fake news detection, dubbed C$^2$EFND. 
Unlike traditional tasks where the model is trained once and directly tested on the test set, our framework addresses the practical need for continual learning in a dynamic news environment, where emergent events require the model to adapt with minimal labeled data while preserving the ability to detect past events. The challenges of continuous emergent fake news detection can be summarized in three main points:
\begin{itemize}
    \item \textbf{Challenge 1}: \textbf{Scarcity of Labeled Data:} Emergent events typically lack sufficient labeled samples, creating significant hurdles for effectively training models.
    \item \textbf{Challenge 2}: \textbf{Continuous Knowledge Updating:} The evolving nature of news necessitates mechanisms for both LLMs and SLMs to update their knowledge without forgetting previously learned information continually.
    \item \textbf{Challenge 3}: \textbf{Effective Collaboration:} Exploiting the complementary strengths of LLMs and SLMs in a collaborative manner is challenging due to their distinct architectures and training paradigms.
\end{itemize}

To address challenge 1, a two-stage active learning framework is proposed. In the first stage, a diversity-based sampling method is applied to identify samples significantly different from previously seen events. In the second stage, an uncertainty-based approach is introduced by leveraging predictions from both LLMs and SLMs for data selection. To address challenge 2, a rehearsal replay mechanism is applied for the continual learning of the SLM, preventing catastrophic forgetting through the periodic replay of previously learned knowledge. Simultaneously, for LLMs, we propose a novel lifelong knowledge editing strategy based on a Mixture-of-Experts (MoE) architecture, enabling efficient incremental updates specific to emergent news events without extensive retraining. To address challenge 3, we develop a multi-round collaborative learning framework that systematically leverages both LLM's and SLM's discriminative ability. This framework employs semi-supervised learning through iterative rounds of clean and noisy data identification, collaboratively refining model predictions and incrementally enhancing the understanding of emergent events.

By addressing these challenges, our proposed C$^2$EFND framework enjoys three notable advantages compared to previous methods as illustrated in Figure~\ref{fig:intro_1}(b).  First, it significantly reduces the dependency on extensive labeled data by strategically combining a two-stage active learning method and iterative multi-round learning. Second, our framework substantially enhances model generalization by enabling seamless collaborative interactions between the comprehensive reasoning abilities of LLMs and the specialized classification skills of SLMs across diverse events. Third, it incorporates robust continual learning mechanisms, including rehearsal-based replay for SLMs and an efficient MoE-based lifelong knowledge editing strategy for LLMs, thus maintaining consistent adaptability to continuously evolving news events. These advances collectively enable C$^2$EFND to achieve SOTA performance in two real-world datasets Twitter16 and Pheme, demonstrating its potential to significantly enhance the accuracy and practicality of fake news detection systems.
Our contributions are summarized as follows:
\begin{itemize}
    \item We formulate a novel and practical fake news detection task under a continual learning paradigm with limited data, and further propose an integrated multi-round collaborative learning framework for joint training of LLMs and SLMs.
    \item We propose a replay-based continue learning mechanism for the SLMs and an MOE-based lifelong knowledge edit mechanism for the LLMs to update their knowledge of continuous emergent news events.
    \item  To effectively leverage scarce manually labeled data, we design a two-stage active learning module combined with an iterative multi-round data selection process to filter clean samples for semi-supervised learning.
    \item Extensive experiments on two real-world datasets Pheme and Twitter16 demonstrate C$^2$EFND achieves SOTA performance in continuous emergent fake news detection.
\end{itemize}

\section{Related Work}
\subsection{Fake News Detection}
Fake news detection is fundamentally a binary classification task. Traditional deep learning-based methods can be divided into three categories: content-based methods, knowledge-enhanced methods, and propagation-enhanced methods~\cite{zhou2020survey}. While content-based methods analyze semantic information~\cite{khattar2019mvae}, they struggle with the evolving nature of fake news~\cite{capuano2023content}. Consequently, recent works incorporate external knowledge and social network  to help detection~\cite{zhang2024reinforced,jing2025dpsg}.

While prior methods perform well on specific events, they often fail to generalize across diverse types of fake news. This has motivated research on multi-domain detection~\cite{silva2021embracing,zhu2022memory,tong2024mmdfnd,liu2023robust}. For instance, M$^3$FEND~\cite{zhu2022memory} models news from a multi-view perspective and proposes a domain memory bank to enrich domain information. MMDFND~\cite{tong2024mmdfnd} incorporates domain embeddings and attention mechanism to effectively utilize information from different domains. 

However, both traditional methods and multi-domain approaches suffer from two major limitations: Firstly, the lack of mechanisms for knowledge updating. As a result, the model requires complete retraining with all data when emergent news events occur, which is highly impractical for real-world applications. Secondly, these methods assume the availability of sufficient labeled data for training, but the cost of annotating fake news is prohibitively high. Furthermore, in emergent situations, there are often limited labeled examples, leading to a sharp decline in the model's detection ability.

\subsection{Continue Learning}
Continue learning refers to the ability of a model to incrementally acquire knowledge over time without forgetting previously learned knowledge, enabling it to adapt to new data and environments~\cite{wang2024comprehensive}. Continue learning can be mainly divided three categories: task-incremental learning~\cite{hyder2022incremental}, class-incremental learning~\cite{belouadah2021comprehensive} and domain-incremental learning~\cite{van2022three}. The setting that is most relevant to our work is domain incremental learning, which aims to train the model to handle the distribution changes of input data.

The majority of methods can be divided into three types: regularization-based approaches, optimization-based approaches, and replay-based approaches~\cite{wang2024comprehensive}. Regularization-based approaches regularize the variation of network parameters to maintain their ability on past data~\cite{li2017learning}. Optimization-based approaches design optimization programs for continue learning~\cite{chaudhry2018efficient}.  The continue learning approach for the SLM in our work closely aligns with the last replay-based approaches~\cite{shi2024unified}, which mitigates catastrophic forgetting by storing a small set of past samples and learning new representations jointly with recent samples.

\subsection{Large Language Models}
Large language models (LLMs) have shown strong performance across NLP tasks~\cite{zhao2023survey,chang2024survey}. However, previous studies have shown that LLMs struggle to assess news veracity in zero-shot settings~\cite{hu2024bad,zhou2025collaborative}. Therefore, rather than utilizing LLMs for inference and classification, more researches have focused on how to utilize LLMs as agents to provide additional knowledge to enhance the detection capability of SLMs~\cite{wan2024dell}. 
For instance, 
Wu et al.~\cite{wu2024fake} employs LLMs to generate trustworthy-style fake news to train a more robust fake news detector. 
ARG leverages the reasoning ability of LLMs to generate decision rationales for SLMs, effectively improving the SLM's detection performance~\cite{hu2024bad}.

Although the aforementioned methods using LLMs effectively enhance the detection performance of SLMs, they have several shortcomings: Firstly, they only utilize frozen-parameters LLMs, which limits the ability of the LLMs to adapt to emergent news environment. Moreover, they overlook the generalization and learning capabilities that LLMs possess due to large-scale pre-training, which could assist SLMs in learning from new samples~\cite{yang2024unveiling}. 
To address the aforementioned issues, our model primarily utilizes two main techniques: knowledge edit and in-context learning~\cite{de2021editing,min2022rethinking}. 

Suffering from outdated or false knowledge usually causes LLMs unaware of unseen events or make predictions with incorrect facts~\cite{martino2023knowledge}. To this end, numerous knowledge edit methods are proposed to update LLMs with the latest and accurate knowledge~\cite{de2021editing}. Initially, knowledge edit focuses on single or batch edits, overlooking the need for continuous updates in real-world scenarios. As a result, the concept of lifelong knowledge edit is introduced, enabling the incorporation of continuously evolving knowledge from data streams into LLMs~\cite{hartvigsen2024aging}.
In our method, we propose a novel machine for lifelong knowledge edit LLMs to help them understand emergent news events while remaining detection abilities with past events.
As model and data sizes increase, LLMs exhibit in-context learning capabilities, which enable them to make more precise predictions based on a few demonstrations provided within the context~\cite{dong2022survey}. In our work, we utilize in-context learning to help LLMs understand the task of fake news detection and augment their ability in reasoning and detecting  by providing suitable demonstrations.

\section{Problem Formulation}
This paper primarily aims to address the continue learning of emergent fake news detection. Let $\epsilon$ denote a set of news events while each event $e$ is accompanied by its timestamp. The news events $\epsilon$ is sorted by these timestamps, arranged from earliest to latest: $\epsilon = (e_1,t_1),(e_2,t_2),...,(e_n,t_n)$. The model will be trained and tested sequentially in timestamp order. In each event $e_i$, only a limited number of samples are annotated: $\{\mathcal{X}_i,\mathcal{Y}_i\} = \{x_i,y_i\}_{i=1}^{\hat{N}_{e_i}}$, where $\hat{N}_{e_i} \textless {N}_{e_i}$. The remaining unlabeled samples are defined as the testset: $ \mathcal{X}_i = \{x_i\}_{i=\hat{N}_{e_i}}^{N_{e_i}}$. 
A memory bank is denoted as $\mathcal{M} = \{\mathcal{M}_1,\mathcal{M}_2,...,\mathcal{M}_n\}$, each $\mathcal{M}_i \in \mathcal{M}$ denotes the memory bank which consists of labeled samples of event $e_i$: $\mathcal{M}_i = \{\{\mathcal{X}_i,\mathcal{Y}_i\} = \{x_i,y_i\}_{i=1}^{N_{e_i}}, x_i \in e_i \}$.
The final objective is for the LLM $\mathcal{L}$ and the SLM $\mathcal{S}$ to continually detect the authenticity of emergent news events.

\section{Methodology}
\subsection{Active Learning for Annotation}
Emergent news events typically arrive without labels and manual annotation is both costly and time-consuming~\cite{zhou2020survey}
Therefore, we introduce an approach based on deep active learning to select a small but informative subset of data for efficient event understanding~\cite{li2024survey}. 

Given the unlabeled news $ \mathcal{X}_i=\{x_i\}_{i=1}^{N_{e_i}}$  of event $e_i$ and existed labeled news from memory bank $\mathcal{M} = \{\mathcal{M}_1,\mathcal{M}_2,...,\mathcal{M}_{i-1}\}$. As external knowledge contributes significantly to fake news detection models~\cite{zhang2024reinforced}, we first extract the relevant knowledge of key entities for each news item $x_i$ from Wikipedia. A LLM is utilized as an agent to retrieve key entities $\{k_1,k_2,...,k_n\}$ from $x_i$  and the Wikipedia API is used to retrieve latest and accurate information about these key entities: $k_{x_i} = \{(k_1,i_1),(k_2,i_2),...,(k_n,i_n)\}$. All entities of $\mathcal{X}_i$ is then integrated as $\mathcal{K}_i = \{(k_{x_1},k_{x_2},...,k_{x_{N_{e_i}}})\}$

Due to the high time and labor costs of manual annotation, an effective data selection mechanism is required to identify and label only the most essential data.
The proposed human annotation process can be divided into two stages: the first stage is diverse-based selection while the second stage is uncertainty-based selection which is introduced in detail in section~\ref{section:collaborative}. The diversity-based methods aim to sample the most prototypical data points that are different from the labeled samples from $\mathcal{M}$~\cite{hasan2018context}. 

For each unlabeled news $x_i$, we calculate the sum of euclidean distance with existed labeled news $\{x_i,y_i\}_{i=1}^{M}$ from the memory bank. The greater the total distance between a news article and the labeled data in the memory bank, the more it differs from existing labeled data, indicating that it can provide the model with knowledge of emergent news events~\cite{li2024survey}. 
A pretrained RoBERTa~\cite{liu2019roberta} is utilized as an encoder $E$ to convert each news into embedding:
\begin{equation}
\begin{aligned}
    d(x_i,x_j) = \sqrt{(E(x_i)-E(x_j))^2} \\
    s(x_i) = \sum_{j=1}^{m} d(x_i,x_j), x_j \in \mathcal{M} \\
\end{aligned}
\end{equation}
where $d(x_i,x_j)$ denotes the distance between unlabeled news $x_i$ and labeled news $x_j$, $s(x_i)$ denotes the sum of its distance with data in the memory bank $\mathcal{M}$. 
At this stage, the first annotation threshold is set as $\mathcal{P}_1$, which denotes the percentage of data to be selected for labeling. The news $x_i$ with the greatest total distance  will be selected for annotation and added to the memory bank $\mathcal{M}_i$:
\begin{equation}
\begin{aligned}
    \mathcal{M}_{i} = \{ \{x_i,y_i,k_i\} | x_i \in & \mathcal{X}_i, s(x_i) \in max_{\mathcal{P}_1}(s(\mathcal{X}))\} \\
    \mathcal{M} \leftarrow& \mathcal{M} \cup \mathcal{M}_i
\end{aligned}
\end{equation}

When $\mathcal{M} = \emptyset$, i.e., for the first event, we randomly select $\mathcal{P}_1$ proportion of samples from $\mathcal{X}_i$ for initial annotation.

\begin{figure*}[t]
\centering
\includegraphics[scale=1.0]{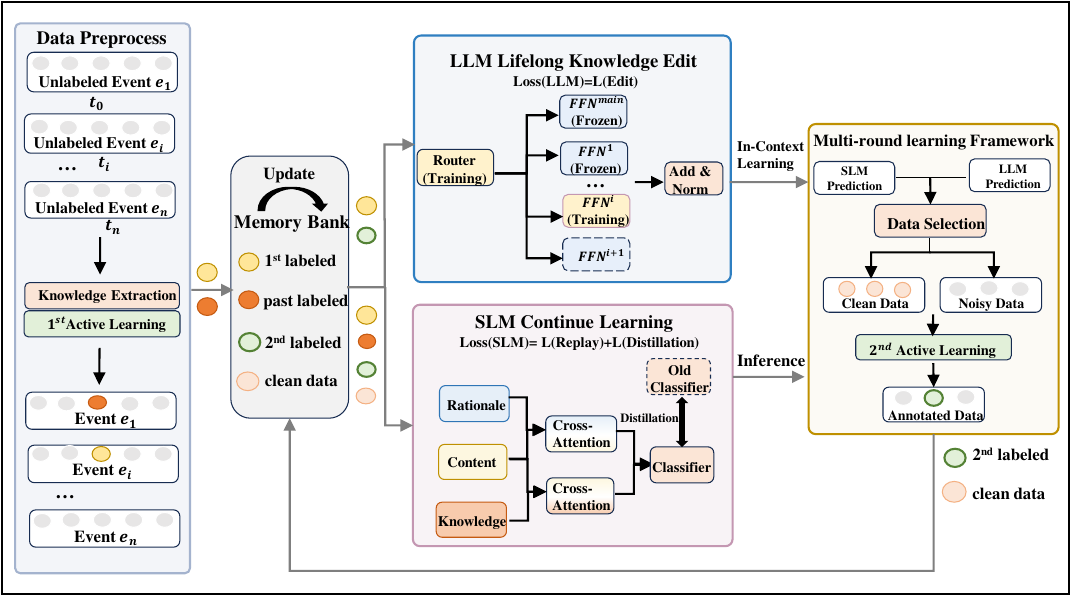}
    \caption{The architecture of C$^2$FEND. C$^2$FEND is composed of four main modules: a data initialization module, a lifelong knowledge editing module for the LLM, a continual learning module for the SLM, and a multi-round collaborative learning module between the LLM and SLM.}
    \label{fig:method}
\end{figure*}
\subsection{Lifelong Knowledge Edit of LLMs}
Due to the high computation cost of fine-tuning LLMs, previous LLM-augmented approaches for fake news detection solely rely on frozen-parameters LLMs, limiting the model's ability to comprehend the latest news events~\cite{wan2024dell}. To address this, we introduce knowledge editing techniques to enable LLMs to incorporate emergent news with only updating few parameters~\cite{de2021editing}. However, traditional knowledge editing methods are typically limited to single or batch updates, overlooking the need for continuous learning in LLMs~\cite{martino2023knowledge}. 
Therefore, we propose a MoE-based lifelong knowledge editing mechanism, enabling LLMs to continuously learn and adapt to evolving emergent news events.

Previous research has demonstrated that fine-tuning the FFN layers (feedforward network) in transformer architectures effectively enables LLMs to learn new knowledge~\cite{de2021editing}. 
Specifically, each FFN module is divided into two MLP layers with an activation function, represented as follows:
\begin{equation}
    f(x) = \sigma(x \cdot \mathcal{W}_{fc}) \cdot \mathcal{W}_{proj}
\end{equation}
where $\mathcal{W}_{fc} \in \mathcal{R}^{d*d_m}$ and $\mathcal{W}_{proj} \in \mathcal{R}^{d_m*d}$ denotes the parameter matrices of the two MLP layers in FFN, $d_m$ denotes the dimension of the intermediate hidden layer in the FFN, $\sigma$ denotes the activation function(e.g. GeLU) and $x \in \mathcal{R}^d$ denotes the input to the FFN. Following previous works, we choose the projection matrix $\mathcal{W}_{proj}$ for knowledge editing.

However, due to the continuous evolution of emergent news events, the knowledge required by the model is constantly updated. Applying multiple knowledge edits to the same FFN layer can lead the model to forget previous edits, focusing solely on current events and resulting in catastrophe forgetting~\cite{hartvigsen2024aging}.
Inspired by the MoE architecture and its successful application in pre-training large language models~\cite{jiang2024mixtral,dou2024loramoe}, we incorporate multiple FFN layers within the transformer to learn knowledge specific to each event respectively.  
Specifically, when a new event $e_i$ occurs, we add and train a dedicated $i_{th}$ FFN layer $FFN^i$ with the labeled data from $\mathcal{M}_i$ and knowledge $\mathcal{K}_i$ while keeping other FFN layers frozen, including the original $FFN_{main}$ module and the added $\sum_{j=1}^{i-1} FFN^j$ module for past event. The transformation of labeled data and extracted knowledge into corresponding input $x$ and output $y$ for knowledge edit will be detailed introduced in Section~\ref{edit_case}.
Similar to common approaches in MoE, a Router Network $\mathcal{R}$ is proposed to compute the proportional score contribution of each expert during inference, which is defined as:
\begin{equation}
    R(x)_i = softmax(\mathcal{W}_g^T \cdot x)
\end{equation}
where $\mathcal{W}_g^T$ denotes the trainable weights of the router network with $d_m$ as the input dimension and $n$ as the number of experts.
The final output of the MOE-FFN network is by selecting the top-k1 experts for inference. The output $h$ of this module can be expressed as:

\begin{equation}
\begin{aligned}
    r(i|x) = & Top_{k1}(R(x)_i) \\
    h(x) = &\frac{f_0(x)+\sum_{j=1}^{i} r(i|x)f_i(x)}{k_1+1}
\end{aligned}
\end{equation}
where $f_i$ denotes the $i^th$ expert of FFN modules, $f_0$ denotes the original FFN module of the LLM $\mathcal{L}$.
The loss function is defined as:

\begin{equation}
\label{llm_loss}
    \mathcal{L}_{edit} = - \sum_{(x,y) \in \mathcal{M} \cup \mathcal{K}} log P(y | x; \theta_{R},\theta_{proj})
\end{equation}

Rationales generated by LLMs through Chain-of-Thought (COT) prompting have been widely used to enhance the reasoning capabilities of SLMs~\cite{hu2024bad}. As the LLM $\mathcal{L}$ has the latest and accurate understanding of emergent news events $e_i$ by knowledge edit, the CoT approach can effectively extract the underlying logic of LLM $\mathcal{L}$ for detecting the authenticity of news. Thus, for each news item $x_i \in e_i$, we utilize the knowledge edited LLM $\mathcal{L}$ to generate rationales
$r_i$, which is then fed into the SLM $\mathcal{S}$ for training. The details of CoT are introduced in Sec~\ref{prompt}.

\subsection{Incremental Learning of SLMs}
Traditional SLM-based methods require retraining from scratch with the entire dataset, which does not align with our setting as emergent events continuously occur. To enable the SLM $\mathcal{S}$ to adapt to emergent events while retaining detection capabilities for past events, a continue learning method is proposed for SLM's incremental learning. Moreover, previous studies have shown that external knowledge and rationales provided by LLMs can support SLMs in making judgments~\cite{hu2024bad,zhang2024reinforced}. These two additional factors are also leveraged to help the SLMs for better assessment.

Specifically, the labeled data from $\mathcal{M}_i$ is utilized to train the SLM with its corresponding knowledge and rationale provided by LLM $\mathcal{L}$, denoted as $x_{train} = \{(x_i,k_i,r_i,y_i)|x_i \in \mathcal{M}_i\}$. Three pre-trained RoBERTa~\cite{liu2019roberta} models are utilized as encoders for the three elements:

\vspace{-0.2cm}
\begin{equation}
    (h_t,h_k,h_r) = RoBERTa(t,k,r), (h_t,h_k,h_r) \in \mathcal{R}^{3 \cdot d1}
\end{equation}

To enhance the model's understanding of news content $x_i$, external knowledge $k_i$, and rationale $r_i$ for better detection, we employ a cross-attention mechanism, which is commonly used in multimodal learning to enable the model to adaptively select useful elements during inference~\cite{wei2020multi}:

\begin{equation}
\begin{aligned}
    h_{k \rightarrow t} = &softmax(\frac{Q_1 \cdot K}{ \sqrt{d1}} ) \cdot V \\
     h_{r \rightarrow t} = &softmax(\frac{Q_2 \cdot K}  {\sqrt{d1}} ) \cdot V \\
     h_{I}& = [h_{k \rightarrow t};h_{r \rightarrow t}]
\end{aligned}
\end{equation}
where $Q_1 = \mathcal{W}_{Q_1} \cdot h_k$, $Q_2 = \mathcal{W}_{Q_2} \cdot h_r$, $K = \mathcal{W}_{K} \cdot h_t$, $V = \mathcal{W}_{V} \cdot h_t$, $d_1$ denotes the dimension of $h_t$ and $[;]$ denotes the concatenation operation. The final representation $h_{I}$ is then fed into a classifier $f_{cls}$ which consists of multiple MLP layers for classification. 
To handle class imbalance within the memory unit $\mathcal{M}_i$, the weighted cross-entropy loss is utilized as the classification loss function:
\begin{equation}
\mathcal{L}_{cls}(y,\hat{y}) = \mathbb{E}_{(x,y) \in \mathcal{M}_i}
\left[-w_1 y \log(\hat{y}) - w_2 (1 - y) \log(1 - \hat{y})\right]
\end{equation}
where $w_1$ and $w_2$ denotes the proportion of real and fake news samples in $\mathcal{M}_i$.

However, training exclusively on labeled data from new events $e_i$ leads to the loss of detection capabilities for past events, resulting in catastrophic forgetting~\cite{van2022three}. To address this, we adopt two continue learning strategies: using data stored in memory bank $\mathcal{M}_{1}^{i-1}$ from past events for rehearsal replay and applying knowledge distillation from the model $\mathcal{S}^{i-1}$ trained from the last event $e_{i-1}$ to prevent complete forgetting of past information~\cite{buzzega2020dark,shi2024unified}. The loss function is as follows:
\begin{equation}
\begin{aligned}
\mathcal{L}_{replay}(y,\hat{y}) = E_{(x,y) \in \mathcal{M}_1^{i-1}} &
 [-y \log(\hat{y})- (1-y) \log(1-\hat{y})] \\
\mathcal{L}_{dil} = E_{x \in \mathcal{M}_1^{i-1}}&[D_{KL}(f_{i-1}(x) || f_{i}(x))]
\end{aligned}
\end{equation}

where $D_{KL}$ denotes the kullback-leibler divergence, $f_i(x)$ denotes the output of current SLM model $\mathcal{S}$ and $f_{i-1}(x)$ denotes the output of last trained SLM model $\mathcal{S}^{i-1}$.
Finally, the overall loss fuction to train SLM $\mathcal{S}$ can be denoted as:
\begin{equation}
\label{slm_loss}
    \mathcal{L}_{slm} = \mathcal{L}_{cls} + \lambda_1 \mathcal{L}_{replay} + \lambda_2 \mathcal{L}_{dil}
\end{equation}

\subsection{Collaborative Multi-Round Learning}
\label{section:collaborative}
Through the aforementioned methods, we have utilized manually labeled data from the first stage annotation for the continue learning of SLM $\mathcal{S}$ and the knowledge editing of LLM $\mathcal{L}$. However, relying on this limited labeled data alone fails to provide a sufficient understanding of emergent events~\cite{silva2021embracing}. Therefore, inspired by semi-supervised learning, we aim to design a framework that effectively leverages unlabeled data to further enhance the model's potential detection ability~\cite{van2020survey}. Moreover, in the previous process, the knowledge-edited LLM $\mathcal{L}$ primarily provided rationales for the SLM $\mathcal{S}$ without fully utilizing its own discriminative capabilities. A collaborative framework is needed to simultaneously leverage the discriminative capabilities of both LLM $\mathcal{L}$ and SLM $\mathcal{S}$.

To this end, a collaborative multi-round learning framework is proposed to fully utilize the unlabeled data and the detection abilities of both the LLM $\mathcal{L}$ and the SLM $\mathcal{S}$ simultaneously. To enhance the detection performance of LLM $\mathcal{L}$, we employ in-context learning and knowledge-augmented methods to retrieve effective demonstrations from the memory bank $\mathcal{M}$. Additionally, a data selection module is designed to identify clean samples from unlabeled data for training, while noisy samples are flagged for the second stage of human annotation. Ultimately, the multi-round learning framework enables both LLM $\mathcal{L}$ and SLM $\mathcal{S}$ to fully leverage unlabeled data and achieve comprehensive understanding of emergent events.

\subsubsection{In-Context Learning of LLM}
In-context learning aims to find suitable demonstrations for LLM $\mathcal{L}$ to understand the news under detection and the task of fake news detection. Previous studies have shown that in-context learning benefits from the similar distribution of test data and demonstration data~\cite{liu2022makes}. To this end, BM25 algorithm is utilized to retrieve most semantically and structurally similar top-k2 data from $\mathcal{M}$ as demonstrations $\mathcal{D}(x)$ for news item $x$ under detection:
\begin{equation}
\begin{aligned}
\mathcal{D}(x) = \{ \sum_{i=1}^{k2} (x_i,y_i) |
(x_i,y_i) \in \text{BM25}_{\text{top-k2}}(x, \mathcal{M}) \}
\end{aligned}
\end{equation}
Given the demonstrations $\mathcal{D}(x)$, external knowledge $k(x)$ extracted by the LLM agent from $\mathcal{K}$ and news item $x$. LLM $\mathcal{L}$ provides the predicted label $\hat{y_1}$ while SLM $\mathcal{S}$ makes the prediction $\hat{y_2}$ of the news with corresponding rationales $r(x)$ and knowledge $k(x)$:

\begin{equation}
\begin{aligned}
\hat{y_1}=argmax_{\hat{y_1} \in \mathcal{Y}}P(\hat{y_1}| \mathcal{D}(x),k(x),x) \\
\hat{y_2}=argmax_{\hat{y_2} \in \mathcal{Y}}P(\hat{y_2}| r(x),k(x),x)
\label{equation_llm}
\end{aligned}
\end{equation}

\begin{algorithm}[!t]
  \SetAlgoNlRelativeSize{-1}
  \SetKwData{Left}{left}\SetKwData{This}{this}\SetKwData{Up}{up}
  \SetKwFunction{Union}{Union}\SetKwFunction{FindCompress}{FindCompress}
  \SetKwInOut{Input}{Input}\SetKwInOut{Output}{output}
  \Input{Emergent Events $ \mathcal{X}_i = \{x_i\}_{i=1}^{N_{e_i}}$, edited LLM $\mathcal{L}^{i-1}$ of last event $e_i$, trained SLM $\mathcal{S}^{i-1}$ of last event $e_i$, memory bank $\mathcal{M}=\{\mathcal{M}_1,\mathcal{M}_2,...,\mathcal{M}_{i-1}\}$, round $=$ 1}
  \Output{Labeled Emergent Events $\{X_e^t\} = \{x_{e,i}^t,y_{e,i}^t\}_{i=K+1}^N$, trained SLM $\mathcal{S}^{i}$, edited LLM $\mathcal{L}^{i}$}
  \tcc{The first Stage Active Learning}\
  \For{each news item $x_i \in \mathcal{X}_i$}{
  Knowledge extraction to obtain $k_{x_i}$\;
  Distance calculation with $\mathcal{M}$ to obtain $s(x_i)$\; 
  }
  Aggregate $k_i$ to obtain event knowledge $\mathcal{K}_i$\; 
  Obtain $\mathcal{M}_i$ by diversity-based active learning with $s(x_i)$, $\mathcal{M} = \mathcal{M} \cup \mathcal{M}_i$\;
  
  \tcc{Lifelong knowledge edit of $\mathcal{L}$}
  Create $FFN^i$ for knowledge edit of event $e_i$\;
  Frozen $\sum_{j=1}^{i-1} FFN^j$ and $FFN_{main}$\;
  Train $FFN^i$ of $\mathcal{L}$ with $\mathcal{K}_i$ and $\mathcal{M}_i$ by Equation~\ref{llm_loss}\;
  
  \tcc{Continue Learning of $\mathcal{S}$}
  Update $\mathcal{S}$ with $\mathcal{S}^{i-1}$ and $(r_i,k_i,x_i,y_i)$ by Equation~\ref{slm_loss}\;
  
  \tcc{Multi-round Learning}\
  \If{round == 1}{
    Obtain Demonstrations $\mathcal{D}$ from $\mathcal{M}$ by BM25\;
    Inference by $\mathcal{L}$ and $\mathcal{S}$ to obtain $(\widehat{y_1},\widehat{y_2})$\;
    $\widehat{y_1}=argmax_{\widehat{y_1} \in \mathcal{Y}}P(\widehat{y_1}| \mathcal{D},\mathcal{K},x)$\;
    $\widehat{y_2}=argmax_{\widehat{y_2} \in \mathcal{Y}}P(\widehat{y_2}|x,k,r)$\;
    Data selection model to obtain  $D_{clean}$ and $D_{noisy}$\;
    $2^{nd}$ active learning to obtain $D_{label}$ and update $\mathcal{M}_i$\;
    round = round + 1\;
  }
  \For {round $\leq$ $\mathcal{N}$}{
    Edit $\mathcal{L}$ with $D_{label}$ and fine-tune $\mathcal{S}$ by $D_{clean} \cup \mathcal{M}$\;
     $D_{noisy}$ inference by $\mathcal{L}$ and $\mathcal{S}$ to obtain $(\widehat{y_1},\widehat{y_2})$\;
     Data selection to update $D_{clean}$ and $D_{noisy}$\;
     $2^{nd}$ active learning to update $D_{label}$ and $\mathcal{M}_i$\;
     round = round + 1\;
  }
  \If{$D_{noisy} \neq \emptyset$}{
    $D_{noisy}$ inference by $\mathcal{S}$ to obtain $\widehat{y_2}$\ as final labels\;
    Update $\mathcal{M}$ by Equation~\ref{memory_update}\;
  }
\caption{Pseudo-code for C$^2$EFND}
\label{alg1}
\end{algorithm}
\subsubsection{Data Selection Module}
\label{data_selection}
In semi-supervised learning, utilizing pseudo labels with high confidence as true labels for further training is a widely used approach~\cite{van2020survey}. Inspired by this, a filtering mechanism that leverages the judgment results of both SLM and LLM to filter unlabeled samples is proposed, classifying all unlabeled data into clean data samples $D_{clean}$ and noisy data samples $D_{noisy}$. Specifically, all unlabeled news $\{(x,\hat{y_1},\hat{y_2}), x \in \mathcal{X}_{test}\}$ are divided into clean data pool $D_{clean}$ and noisy data pool $D_{noisy}$:
\begin{equation}
\begin{aligned}
D_{clean} = \{(x_i,y_i) \ &| \  \hat{y_1} = \hat{y_2} \ and \  p(\hat{y_2}) \geq \omega\} \\
D_{noisy} = \{(x_i) \ &|\  \hat{y_1} \neq \hat{y_2} \ or \  p(\hat{y_2}) < \omega\}
\end{aligned}
\label{equation2}
\end{equation}
where $p(\hat{y_2})$ denotes the output probability by SLM and $\omega$ is a hyper-parameter that denotes the confidence threshold for data selection.

Data in clean data pool $D_{clean}$ represents samples for which LLM and SLM yield consistent detection results with high confidence from SLM. In contrast, data from noisy data pool $D_{noisy}$ have low confidence or conflicting predictions between the LLM and SLM. Following the uncertainty-based active learning method~\cite{nguyen2022measure}, samples in the noisy pool with the lowest confidence undergo the second stage of manual annotation, with newly labeled samples $D_{label}$ added to the memory bank $\mathcal{M}_i$:
\begin{equation}
\begin{aligned}
    D_{label} =  \{ \{x_i,y_i\} | x_i \in  &\mathcal{X}_i, p(x_i) \in min_{\mathcal{P}_2}(s(\mathcal{X}))\} \\
    \mathcal{M}_i &\leftarrow \mathcal{M}_i \cup D_{label}
\end{aligned}
\end{equation}
where $\mathcal{P}_2$ denotes the percentage of data to be selected for labeling in the second annotation stage.

\subsubsection{Multi-Round Learning}
The initial classification of data into $D_{clean}$ and $D_{noisy}$ is only the beginning. To ensure both LLM $\mathcal{L}$ and SLM $\mathcal{S}$ can learn from more data, we use data from clean data pool $D_{clean}$ and updated memory bank $\mathcal{M}$ to finetune the SLM $\mathcal{S}$ based on Equation~\ref{slm_loss}. Since the success of in-context learning relies more on the relevance of demonstration content rather than label accuracy~\cite{liu2022makes}, data with pseudo labels from the clean data pool $D_{clean}$ is also used for in-context learning with the LLM $\mathcal{L}$. Given the high accuracy requirements for knowledge in knowledge editing, only the new manually annotated data from $D_{label}$ are used for another iteration of knowledge edit to the LLM $\mathcal{L}$ based on Equation~\ref{llm_loss}.

After new data is incorporated into the learning processes of both LLM $\mathcal{L}$ and SLM $\mathcal{S}$, another round begins as they re-evaluate the samples from $D_{noisy}$. The noisy data pool $D_{noisy}$ is then reclassified into a new clean data pool $D_{clean}^{new}$ and a new noisy data pool $D_{noisy}^{new}$, also repeating the previous annotation process depicted in Section~\ref{data_selection}. Then the $D_{clean}^{new}$ is integrated with the existing clean data pool $D_{clean}$, and the newly annotated samples $D_{label}^{new}$ selected from $D_{noisy}^{new}$ are added to the memory bank $\mathcal{M}_i \in \mathcal{M}$. This iterative learning continues for multiple rounds until a specified hyper-parameter threshold $\mathcal{N}$ is reached. At this point, any remaining samples in the noisy data pool $D_{noisy}$ are assigned to the clean data pool $D_{clean}$ using the output labels determined by the SLM $\mathcal{S}$, completing the detection and learning for this event $e_i$. 
The overall pipeline is depicted in Algorithm~\ref{alg1}, providing a detailed visualization of the multi-round learning process.

\subsubsection{Memory Bank}
To ensure the label quality within the memory bank, we first remove pseudo-labeled samples from the clean pool once the multi-round learning process is completed: $\mathcal{M}_i \leftarrow \mathcal{M}_i \setminus \mathcal{D}_{clean}$. 

Due to limited storage capacity and the design objective that the model should not rely on learning from all labeled samples, the memory bank is constrained by a maximum storage threshold denoted as \(M_{max}\). 
When the size of the Memory Bank exceeds the predefined maximum capacity \(\vert M \vert > M_{max}\), we adopt a diversity-based clustering approach to ensure optimal representation~\cite{zou2025glean}. Specifically, each data point \(x_j \in M\) is encoded by a pretrained RoBERTa model to obtain its semantic embeddings \(h_j = E(x_j)\). Subsequently, we perform k-means clustering on these embeddings, setting the number of clusters equal to \(M_{max}\). We retain the representative data point \(x_k^*\) closest to each cluster centroid \(c_k\), thus maintaining a diverse subset of historical labeled data:
\begin{equation}
M \leftarrow \{ x_k^* \mid x_k^* = \arg\min_{x_j \in C_k}\|E(x_j)-c_k\|,\; k=1,\dots,M_{max} \}
\label{memory_update}
\end{equation}
where \(C_k\) represents the set of data assigned to the \(k^{th}\) cluster.

This Memory Bank updating strategy ensures controlled storage size, sustained diversity, and effective prevention of catastrophic forgetting during continual learning.

\section{Experiments}
\subsection{Datasets}
To evaluate the proposed method C$^2$EFND, extensive experiments are conducted on two datasets from real-world social media sources, namely Twitter16~\cite{boididou2018detection} and Pheme~\cite{zubiaga2017exploiting}. Both Twitter16 and Pheme contain news articles directly collected from trending events on Twitter, with each piece labeled by its associated event and timestamp, which suits our continuous emergent fake news detection setting well.

The Pheme dataset contains five real news events, which we use for continue learning tests in time order as follows: \textit{Ferguson Unrest}, \textit{Ottawa Shooting}, \textit{Sydney Siege}, \textit{Charlie Hebdo Shooting}, \textit{Germanwings Plane Crash}. The Twitter16 dataset also consists of multiple events with timestamps. We consolidate multiple events in the dataset into five time-period-based events in chronological order. The detailed statistical information of these two datasets is in Table ~\ref{pheme_statistic} and Table~\ref{twitter16_statistic}.
\begin{table}[htbp]
    \vspace{-2mm}
  \centering
  \caption{Statistics of the Pheme dataset.}
  \resizebox{1.0\linewidth}{!}{
    \begin{tabular}{l|ccccc}
    \toprule
          & \multicolumn{1}{c}{E1} & \multicolumn{1}{c}{E2} & \multicolumn{1}{c}{E3} & \multicolumn{1}{c}{E4} & \multicolumn{1}{c}{E5} \\
    \midrule
    \multirow{2}[1]{*}{Event} & \multirow{2}[1]{*}{\makecell[c]{Ferguson \\ Unrest}} & \multirow{2}[1]{*}{\makecell[c]{Ottawa \\ Shooting}} & \multirow{2}[1]{*}{\makecell[c]{Sydney \\ Siege}} & \multirow{2}[1]{*}{\makecell[c]{Charlie Hebdo \\ Shooting}} & \multirow{2}[1]{*}{\makecell[c]{Germanwings \\ Plane Crash}} \\
    & & & & &\\
    \midrule
    Real  & 859   & 420   & 699   & 1621  & 238 \\
    Fake  & 254   & 470   & 512   & 458   & 231 \\
    \midrule
    All   & 1143  & 890   & 1221  & 2079  & 469 \\
    \bottomrule
    \end{tabular}%
    }
  \label{pheme_statistic}%
  \vspace{-6mm}
\end{table}%

\begin{table}[htbp]
  \centering
  \caption{Statistics of the Twitter16 dataset.}
  \resizebox{1.0\linewidth}{!}{
    \begin{tabular}{l|ccccc}
    \toprule
          & \multicolumn{1}{c}{E1} & \multicolumn{1}{c}{E2} & \multicolumn{1}{c}{E3} & \multicolumn{1}{c}{E4} & \multicolumn{1}{c}{E5} \\
    \midrule
    Event & \multicolumn{1}{l}{Sandy} & \multicolumn{1}{l}{Boston+Attacks} & \multicolumn{1}{l}{Syrianboy+Eclipse} & \multicolumn{1}{l}{Napal} & \multicolumn{1}{l}{Others} \\
    \midrule
    Real  & 3638  & 726   & 106   & 966   & 354 \\
    Fake  & 3790  & 165   & 638   & 317   & 1775 \\
    \midrule
    All   & 7428  & 891   & 744   & 1283  & 2129 \\
    \bottomrule
    \end{tabular}%
    }
  \label{twitter16_statistic}%
  \vspace{-4mm}
\end{table}%

\subsection{Baselines}
We divide fake news detection methods into three categories: traditional SLM methods, LLM+SLM methods, and multi-domain methods.

\noindent{\textbf{For the single-event SLM methods.}}
\noindent \textbf{RoBERTa}~\cite{liu2019roberta} is an extension of BERT~\cite{devlin2018bert} which employs dynamic masking strategies and larger batch sizes during pre-training for natural language understanding. \textbf{MVAE}~\cite{khattar2019mvae} comprises three components: an encoder to encode the shared representation of features, a decoder to reconstruct the representation, and a detector to classify the truth of posts. \textbf{CompNet}~\cite{hu2021compare} constructs a directed heterogeneous document graph to utilize knowledge base. \textbf{FTT}~\cite{hu2023learn} adapts the model to future data by forecasting the temporal distribution patterns of news data.




\noindent{\textbf{For the SLM+LLM Methods.}}
\noindent \textbf{ARG}~\cite{hu2024bad} designs an adaptive rationale guidance network to help SLM detection with LLM's rationales. \textbf{EFND}~\cite{wang2024explainable} utilizes LLMs to reason about the authenticity of evidences retrieved from news to enhance SLM's ability in fake news detection.

\noindent{\textbf{For the Multi-Domain Method.}}
\noindent \textbf{EANN}~\cite{wang2018eann} firstly utilizes a discriminator to derive event-invariant features for multi-domain detection.
\textbf{MDFEND}~\cite{nan2021mdfend}  leverages a domain gate to aggregate  representations extracted by a mixture of experts for multi-domain fake news detection.  
\textbf{M$^{3}$FEND}~\cite{zhu2022memory} proposes a memory-guided multi-view framework to address the problem of domain shift and domain labeling incompleteness. 
\textbf{CANMD}~\cite{yue2022contrastive} proposes a contrastive adaptation network to solve the label shift problem in fake news detection.

\subsection{Implementation Details}
\subsubsection{Prompt of LLMs}
\label{prompt}
\begin{figure}[h]
\centering
\includegraphics[scale=1.0]{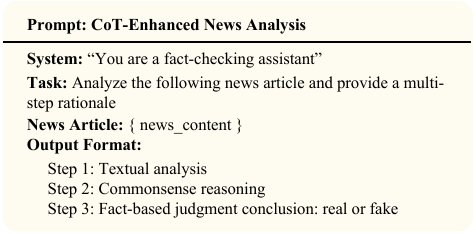}
    \setlength{\abovecaptionskip}{-2mm}
    \caption{Prompt for the LLM to generate rationales.}
    \label{fig:prompt}
\end{figure}

In our framework, we incorporate Chain-of-Thought (CoT) prompting to enhance the interpretability and reasoning capacity of the LLM when assessing the veracity of news content. Drawing inspiration from prior work~\cite{hu2024bad}, we prompt the LLM to generate rationales by considering multiple reasoning dimensions such as emotional tone, commonsense plausibility, and factual consistency. The prompt is illustrated in Figure~\ref{fig:prompt}.

\vspace{-2mm}


\begin{table*}[t]
  \centering
  \caption{Performance of methods on \textbf{target} events on \textbf{Pheme}. Best results are in \textbf{bold} and second best results are \underline{underlined}.}
  \vspace{-2mm}
  \resizebox{1.0\linewidth}{!}{
    \begin{tabular}{c|l|rr|rr|rr|rr|rr}
    \toprule
    \multicolumn{1}{c|}{\multirow{2}[4]{*}{Category}} & \multicolumn{1}{c|}{\multirow{2}[4]{*}{Method}} & \multicolumn{2}{c|}{E1} & \multicolumn{2}{c|}{E2} & \multicolumn{2}{c|}{E3} & \multicolumn{2}{c|}{E4} & \multicolumn{2}{c}{E5} \\
\cmidrule{3-12}          &       & \multicolumn{1}{c}{Accuracy} & \multicolumn{1}{c|}{F1-Score} & \multicolumn{1}{c}{Accuracy} & \multicolumn{1}{c|}{F1-Score} & \multicolumn{1}{c}{Accuracy} & \multicolumn{1}{c|}{F1-Score} & \multicolumn{1}{c}{Accuracy} & \multicolumn{1}{c|}{F1-Score} & \multicolumn{1}{c}{Accuracy} & \multicolumn{1}{c}{F1-Score} \\
    \midrule
    \multirow{4}[2]{*}{\makecell[c]{Single-Event \\ SLM}} & RoBERTa &75.8  $\pm$  0.4      & 86.3  $\pm$  0.7      & 78.2  $\pm$  1.7      &  78.6  $\pm$  0.8     & 77.4  $\pm$  0.3      & 79.3  $\pm$  0.9      & 81.0  $\pm$  0.2      &  87.9  $\pm$  0.2     & 68.8  $\pm$  2.7      & 58.9  $\pm$  6.5 \\
          & MVAE  & 77.3  $\pm$  1.1      & 84.2  $\pm$  1.0      &78.6  $\pm$  0.9       & 78.9  $\pm$  1.0      & 76.5  $\pm$  1.3      & 78.1  $\pm$  1.1      &80.1  $\pm$  1.0       &85.0  $\pm$  1.0       &77.4  $\pm$  1.5       &77.9  $\pm$  1.3  \\
          & CompNet  &78.9  $\pm$  0.8       &85.5  $\pm$  0.9       &79.1  $\pm$  0.7       &79.6  $\pm$  0.6       &77.0  $\pm$  1.0       &79.2  $\pm$  1.2       &81.5  $\pm$  0.6       &86.3  $\pm$  0.8       &78.0  $\pm$  1.4       &78.8  $\pm$  1.5  \\
          & FTT   & 76.8 $\pm$ 1.0 & 83.9 $\pm$ 0.9 & 78.0 $\pm$ 1.2 & 78.4 $\pm$ 0.8 & 75.9 $\pm$ 1.0 & 77.5 $\pm$ 1.1 & 80.4 $\pm$ 0.9 & 84.5 $\pm$ 1.1 & 76.2 $\pm$ 1.3 & 76.8 $\pm$ 1.6 \\
    \midrule
    \multicolumn{1}{c|}{\multirow{2}[2]{*}{LLM+SLM}} & ARG   & 76.2   $\pm$  1.2    & 82.4  $\pm$  0.8      &  80.8  $\pm$  0.9     &  80.7  $\pm$  0.9     & 78.4  $\pm$  0.6      &78.1  $\pm$  0.7       & \underline{83.9  $\pm$  0.4}     & 75.2  $\pm$  2.1      &  81.6  $\pm$  1.6     &81.6  $\pm$  1.6  \\
          & EFND & 75.6 $\pm$ 0.8 & 82.8 $\pm$ 1.0 & 81.2 $\pm$ 0.6 & 81.0 $\pm$ 0.6 & 79.4 $\pm$ 0.3 & 79.2 $\pm$ 0.3 & 83.1 $\pm$ 0.2 & 77.0 $\pm$ 0.3 & 81.2 $\pm$ 0.7 & 81.1 $\pm$ 0.7   \\
    \midrule
    \multirow{4}[2]{*}{\makecell[c]{Multi-Domain \\ SLM}} & EANN  & 81.0  $\pm$  1.2   &  85.6  $\pm$  0.5     & 80.9  $\pm$  0.8    & 81.0  $\pm$  0.7      &   78.3  $\pm$  0.8   &   \underline{81.4  $\pm$  0.2}    & 82.8  $\pm$  0.6     &   89.1  $\pm$  0.2    & 80.4  $\pm$  1.8      & 81.5  $\pm$  0.9 \\
          & MDFEND & 81.7  $\pm$  0.9      &   87.4  $\pm$  0.6   &  82.0  $\pm$  0.8     & \underline{82.0  $\pm$  1.0}      & 78.8  $\pm$  0.5     &   77.5  $\pm$  0.4    &   83.4  $\pm$  0.7   &   88.5  $\pm$  0.4   &  \textbf{82.4  $\pm$  1.4}     & \textbf{82.6  $\pm$  1.4} \\
          & M3FEND & \underline{82.2  $\pm$  0.9}      & \underline{87.8  $\pm$  0.9}      & 82.3  $\pm$  0.7      & 81.8  $\pm$  0.7      & 78.2  $\pm$  1.1      &  81.3  $\pm$  0.4     &  82.9  $\pm$  1.4    & 89.6  $\pm$  0.7      &  78.6  $\pm$  3.0     &79.1  $\pm$  2.3  \\
          & CANMD &80.7  $\pm$  3.9       & 81.3  $\pm$  3.2      & \textbf{82.5  $\pm$  1.1}      &\textbf{82.4  $\pm$  1.2}       &  \underline{79.5  $\pm$  1.1}   &  79.5  $\pm$  1.2     &83.0  $\pm$  1.4       & \underline{89.6  $\pm$  0.7}      &  80.1  $\pm$  1.3    &79.7  $\pm$  1.6  \\
    \midrule
    Ours  & C$^2$EFND   & \textbf{84.4 $\pm$ 1.2} & \textbf{89.7 $\pm$ 1.5} & \underline{82.3 $\pm$ 0.7}& 81.8 $\pm$ 0.6 & \textbf{85.6 $\pm$ 0.8 }& \textbf{82.4 $\pm$ 0.7 }& \textbf{87.2 $\pm$ 0.9} & \textbf{91.5 $\pm$ 0.6} & \underline{82.2 $\pm$ 1.1} & \underline{82.5 $\pm$ 1.3} \\
    \bottomrule
    \end{tabular}%
    }
  \label{pheme_target}%
  \vspace{-4mm}
\end{table*}%

\begin{table*}[t]
  \centering
  \caption{Performance of methods on \textbf{target} events on \textbf{Twitter16}. Best results are in \textbf{bold} and second best results are \underline{underlined}.}
  \vspace{-2mm}
  \resizebox{1.0\linewidth}{!}{
    \begin{tabular}{c|l|rr|rr|rr|rr|rr}
    \toprule
    \multicolumn{1}{c|}{\multirow{2}[4]{*}{Category}} & \multicolumn{1}{c|}{\multirow{2}[4]{*}{Method}} & \multicolumn{2}{c|}{E1} & \multicolumn{2}{c|}{E2} & \multicolumn{2}{c|}{E3} & \multicolumn{2}{c|}{E4} & \multicolumn{2}{c}{E5} \\
\cmidrule{3-12}          &       & \multicolumn{1}{c}{Accuracy} & \multicolumn{1}{c|}{F1-Score} & \multicolumn{1}{c}{Accuracy} & \multicolumn{1}{c|}{F1-Score} & \multicolumn{1}{c}{Accuracy} & \multicolumn{1}{c|}{F1-Score} & \multicolumn{1}{c}{Accuracy} & \multicolumn{1}{c|}{F1-Score} & \multicolumn{1}{c}{Accuracy} & \multicolumn{1}{c}{F1-Score} \\
    \midrule
    \multirow{4}[2]{*}{\makecell[c]{Single-Event \\ SLM}} & RoBERTa & 73.5 $\pm$ 0.4      &   69.6 $\pm$ 1.4     &   78.4 $\pm$ 1.7    & 86.8 $\pm$ 1.4     & 86.5 $\pm$ 0.9      &   69.8 $\pm$ 5.9     &   89.1 $\pm$ 0.7    &  90.9 $\pm$ 0.4    & 79.3 $\pm$ 0.3      & 70.8 $\pm$ 0.9 \\
          & MVAE  & 76.8 $\pm$ 0.5 & 71.2 $\pm$ 1.2 & 80.2 $\pm$ 1.5 & 84.1 $\pm$ 1.3 & 85.3 $\pm$ 1.1 & 77.0 $\pm$ 5.0 & 90.4 $\pm$ 1.0 & 93.1 $\pm$ 0.6 & 80.5 $\pm$ 0.7 & 69.2 $\pm$ 1.4 \\

         & CompNet  & 78.3 $\pm$ 0.6 & 73.4 $\pm$ 1.1 & 81.5 $\pm$ 1.3 & 85.2 $\pm$ 1.0 & 86.0 $\pm$ 0.8 & 78.6 $\pm$ 4.7 & 91.2 $\pm$ 0.9 & \underline{93.9 $\pm$ 0.5} & 82.6 $\pm$ 0.5 & 74.4 $\pm$ 1.2 \\
         & FTT   & 77.4 $\pm$ 0.4 & 72.0 $\pm$ 1.0 & 80.9 $\pm$ 1.1 & 84.5 $\pm$ 1.2 & 85.6 $\pm$ 1.0 & 73.8 $\pm$ 5.2 & 90.7 $\pm$ 0.8 & 93.6 $\pm$ 0.7 & 84.8 $\pm$ 0.6 & 71.9 $\pm$ 1.1 \\
    \midrule
    \multicolumn{1}{c|}{\multirow{2}[2]{*}{LLM+SLM}} & ARG   &\underline{85.0 $\pm$ 0.7}      & \underline{85.0 $\pm$ 0.7}      &   \underline{87.9 $\pm$ 1.6}    &   77.9 $\pm$ 0.4    &  \underline{93.3 $\pm$ 0.8}  &   85.6 $\pm$ 1.7    &   90.5 $\pm$ 0.3   &  91.1 $\pm$ 0.5      &  90.6 $\pm$ 0.2    & 80.6 $\pm$ 0.8  \\
          & EFND   & 83.9 $\pm$ 0.2       &  83.8 $\pm$ 0.2    & 87.1 $\pm$ 0.2     &  81.4 $\pm$ 0.4      &  92.0 $\pm$ 0.3     & 82.6 $\pm$ 0.5      &    \underline{91.9 $\pm$ 0.4}   &  88.9 $\pm$ 0.5     & 89.4 $\pm$ 0.6     &  79.3 $\pm$  1.1\\
    \midrule
    \multirow{4}[2]{*}{\makecell[c]{Multi-Domain \\ SLM}} & EANN   & 82.6 $\pm$ 0.6 & 82.4 $\pm$ 0.4 & 86.4 $\pm$ 0.5 & 90.9 $\pm$ 0.3 & 91.2 $\pm$ 0.5 & 80.2 $\pm$ 1.3 & 91.9 $\pm$ 0.4 & 93.8 $\pm$ 0.2 & 88.2 $\pm$ 0.7 & 68.3 $\pm$ 1.4 \\
& MDFEND & 81.4 $\pm$ 0.4 & 82.6 $\pm$ 0.4 & 87.4 $\pm$ 0.4 & \underline{92.3 $\pm$ 0.2} & 92.8 $\pm$ 0.6 & 79.7 $\pm$ 1.2 & 91.1 $\pm$ 0.6 & 92.0 $\pm$ 0.3 & 87.0 $\pm$ 1.0 & 67.0 $\pm$ 1.1 \\
& M3FEND & 80.4 $\pm$ 1.7 & 80.2 $\pm$ 2.0 & 85.5 $\pm$ 1.7 & 91.5 $\pm$ 0.8 & 93.0 $\pm$ 1.2 & 77.0 $\pm$ 2.5 & 91.0 $\pm$ 0.2 & 91.9 $\pm$ 0.1 & 84.8 $\pm$ 2.0 & 63.4 $\pm$ 4.2 \\
& CANMD  & 84.7 $\pm$ 1.0 & 84.7 $\pm$ 1.0 & 87.1 $\pm$ 1.6 & 87.1 $\pm$ 1.5 & 92.3 $\pm$ 1.1 & \textbf{89.8 $\pm$ 1.5} & 91.6 $\pm$ 1.2 & 91.6 $\pm$ 1.4 & \underline{90.7 $\pm$ 1.3} & \textbf{90.7 $\pm$ 1.3} \\
    \midrule
    Ours  & C$^2$EFND   & \textbf{87.7  $\pm$  1.2}     & \textbf{86.7  $\pm$  1.0}     & \textbf{89.1  $\pm$  1.2}      &  \textbf{93.6  $\pm$  0.7}    & \textbf{94.2 $\pm$ 1.8}     &  \underline{86.5 $\pm$ 2.1}    & \textbf{92.6 $\pm$ 0.6}     & \textbf{94.7 $\pm$ 0.8}      & \textbf{93.8 $\pm$ 0.9}     & \underline{82.1 $\pm$ 1.2} \\
    \bottomrule
    \end{tabular}%
    }
  \label{twitter_target}%
  \vspace{-4mm}
\end{table*}%

\subsubsection{Knowledge Edit Examples}
\label{edit_case}
To enhance the large model’s understanding of news content and improve its discriminative capabilities, we perform knowledge editing using two types of data: factual knowledge extracted from news and news items annotated with veracity labels. The prompt templates for knowledge edit is shown in Table~\ref{tab:edit_prompts}.

\begin{table}[htbp]
  \centering
  \caption{Prompt templates used in knowledge editing scenarios.}
  \vspace{-2mm}
  \renewcommand{\arraystretch}{1.3}
  \begin{tabular}{>{\raggedright\arraybackslash}p{1.2cm} | >{\raggedright\arraybackslash}p{3.2cm} >{\raggedright\arraybackslash}p{1.2cm} >{\raggedright\arraybackslash}p{1.5cm}}
    \toprule
    \textbf{Edit Type} & \textbf{System Prompt} & \textbf{Input} & \textbf{Assistant} \\
    \midrule
    Factual Knowledge & 
    You are a knowledge assistant. Your task is to update your knowledge with new information. & 
    Entity: \{entity\} \{description\} & 
    Understood. \{entity\} is \{description\}. \\
    \midrule
    News Item & 
    You are a fact-checking assistant. Your task is to determine whether a given news article is real or fake. & 
    \{news\} & 
    This is a \{label\} news. \\
    \bottomrule
  \end{tabular}
  \label{tab:edit_prompts}
  \vspace{-4mm}
\end{table}

\subsubsection{Training Implementation}
To align with the low-resource setting of emerging news events as defined in our study, we allocate 20\% of each event-specific dataset for training and validation, while the remaining 80\% is reserved for testing.  For the training of SLM, we set the confidence threshold $\omega$ to 0.85, batch size to 32, round threshold $\mathcal{N}$ to 3, the proportion $\mathcal{P}_1$ of $1^{st}$ active learning is set to 10\% while $\mathcal{P}_2$ of $2^{nd}$ is set to 5\%. The number $k_1$ of expert selection is set to 2 and the number $k_2$ of demonstration selection is set to 8. The storage threshold of memory bank $\mathcal{M}_{max}$ is set to 400. For SLM's fine-tuning, we use the AdamW with a weight decay of 1e-3 as the optimizer and the initial learning rate is set to 1e-4. The weight $\lambda_1$ of replay loss $\mathcal{L}_{replay}$ is set to 1.0 and the weight $\lambda_2$ of distillation loss $\mathcal{L}_{dil}$ is set to 0.4.
For LLM's knowledge edit, We use Llama3-8B Instruct as the LLM $\mathcal{L}$. The initial learning rate is set to 1e-5, the edited layer is set to 26. All experiments are conducted on 4 NVIDIA 3090 GPUs.

\subsection{Experimental Results}
\subsubsection{Results on Target Events}
We first evaluate C$^2$EFND on \textit{target} (unseen emergent) events using the Pheme and Twitter16 datasets, and the results are shown in  Table~\ref{pheme_target} and Table~\ref{twitter_target}. Overall, our approach achieves the highest accuracy and f1-score on most target event in both datasets, demonstrating a superior generalization to new events. We attribute this to our framework’s ability to leverage the latest knowledge and labeled samples for knowledge edit to enhance LLM’s contextual reasoning ability to quickly grasp the new event’s nuances. Traditional single-event methods, lacking any cross-event knowledge transfer, perform markedly worse than other methods. 
Although the LLM+SLM approaches like ARG~\cite{hu2024bad} show improved performance over only utilizing SLMs by leveraging LLMs to provide rationales or explanations, their performance is constrained by the static nature of fixed-parameter LLMs, which hinders the incorporation of updated reasoning information. Multi-domain methods show strong performance among the baselines, since their event-invariant feature learning offers some resilience to domain shift. However, these models still exhibit a noticeable performance gap compared to C$2$EFND. First, they fail to effectively leverage LLMs as auxiliary support. Second, in scenarios with sparse data, the lack of appropriate data selection strategies results in insufficient training of the SLM.

\begin{table*}[t]
  \centering
  \caption{Performance of methods on \textbf{all} events on \textbf{Pheme}. Best results are in \textbf{bold} and second best results are \underline{underlined}.}
    \vspace{-2mm}
  \resizebox{1.0\linewidth}{!}{
    \begin{tabular}{c|l|rr|rr|rr|rr|rr}
    \toprule
    \multicolumn{1}{c|}{\multirow{2}[4]{*}{Category}} & \multicolumn{1}{c|}{\multirow{2}[4]{*}{Method}} & \multicolumn{2}{c|}{E1} & \multicolumn{2}{c|}{E2} & \multicolumn{2}{c|}{E3} & \multicolumn{2}{c|}{E4} & \multicolumn{2}{c}{E5} \\
\cmidrule{3-12}          &       & \multicolumn{1}{c}{Accuracy} & \multicolumn{1}{c|}{F1-Score} & \multicolumn{1}{c}{Accuracy} & \multicolumn{1}{c|}{F1-Score} & \multicolumn{1}{c}{Accuracy} & \multicolumn{1}{c|}{F1-Score} & \multicolumn{1}{c}{Accuracy} & \multicolumn{1}{c|}{F1-Score} & \multicolumn{1}{c}{Accuracy} & \multicolumn{1}{c}{F1-Score} \\
    \midrule
    \multirow{4}[2]{*}{\makecell[c]{Single-Event \\ SLM}} & RoBERTa &75.8 $\pm$ 0.0       & 86.3 $\pm$ 0.0      & 76.4 $\pm$ 1.4      & 70.8 $\pm$ 2.3      & 76.8  $\pm$  1.4      & 80.7  $\pm$  2.3      &  79.1  $\pm$  0.3     &  84.8  $\pm$  0.1     & 78.2  $\pm$  0.8      &83.4  $\pm$  1.0  \\
          & MVAE    & 78.1 $\pm$ 0.5 & 88.1 $\pm$ 0.6 & 78.9 $\pm$ 1.2 & 72.3 $\pm$ 2.1 & 79.4 $\pm$ 1.1 & 82.3 $\pm$ 1.8 & 81.0 $\pm$ 0.4 & 85.2 $\pm$ 0.3 & 80.1 $\pm$ 0.7 & 84.2 $\pm$ 0.9 \\
      & CompNet    & 77.4 $\pm$ 0.8 & 87.2 $\pm$ 0.7 & 77.9 $\pm$ 1.3 & 71.7 $\pm$ 2.2 & 78.3 $\pm$ 1.4 & 81.7 $\pm$ 2.4 & 80.3 $\pm$ 0.5 & 84.3 $\pm$ 0.4 & 79.5 $\pm$ 0.8 & 83.5 $\pm$ 1.1 \\
      & FTT     & 79.2 $\pm$ 0.6 &  \underline{88.7 $\pm$ 0.8} & 79.8 $\pm$ 1.1 & 73.5 $\pm$ 1.9 & 80.0 $\pm$ 1.2 & 82.7 $\pm$ 2.0 & 82.0 $\pm$ 0.4 & 85.7 $\pm$ 0.2 & 81.2 $\pm$ 0.6 & 84.7 $\pm$ 0.8 \\
    \midrule
\multicolumn{1}{c|}{\multirow{2}[2]{*}{LLM+SLM}} & ARG   &76.2  $\pm$ 1.2       & 82.4 $\pm$ 0.8       & 80.2 $\pm$ 0.9      &78.2 $\pm$ 1.4       & 79.5 $\pm$ 1.2      &  78.4 $\pm$ 1.3     &  81.2 $\pm$ 1.0     &  78.0 $\pm$ 1.4     & \textbf{83.2 $\pm$ 0.6}       & 81.4 $\pm$ 0.6 \\
          & EFND   &  75.6  $\pm$  0.8     & 82.8  $\pm$  1.0      & 79.9 $\pm$ 0.6      & 79.3 $\pm$ 0.3      &  80.2 $\pm$ 0.2     & 79.3 $\pm$ 0.2      &  \underline{82.2 $\pm$ 0.2}      &  80.0 $\pm$ 0.2     &   82.1 $\pm$ 0.1    &80.0 $\pm$ 0.2  \\
    \midrule
    \multirow{4}[2]{*}{\makecell[c]{Multi-Domain \\ SLM}} 
          & EANN   & 81.0 $\pm$ 1.2 & 85.6 $\pm$ 0.5 & 80.8 $\pm$ 0.9 & 85.5 $\pm$ 0.4 & 80.0 $\pm$ 0.8 & 84.2 $\pm$ 0.5 & 81.5 $\pm$ 0.5 &  \underline{86.6 $\pm$ 0.2} & 81.0 $\pm$ 1.0 & 86.0 $\pm$ 0.4 \\
& MDFEND & 81.7 $\pm$ 0.9 & 87.4 $\pm$ 0.6 &  \underline{80.9 $\pm$ 0.8} &  \underline{85.6 $\pm$ 0.4} & 80.3 $\pm$ 0.6 &  \underline{84.3 $\pm$ 0.2} & 80.8 $\pm$ 1.4 & 81.5 $\pm$ 0.5 & 82.0 $\pm$ 0.3 &  \underline{86.8 $\pm$ 0.1} \\
& M3FEND &  \underline{82.2 $\pm$ 0.9} & 87.8 $\pm$ 0.9 & 78.5 $\pm$ 0.2 & 83.0 $\pm$ 0.2 & 79.6 $\pm$ 0.8 & 83.7 $\pm$ 0.5 & 80.8 $\pm$ 0.9 & 86.1 $\pm$ 0.5 & 81.0 $\pm$ 0.6 & 85.5 $\pm$ 0.5 \\
& CANMD  & 80.7 $\pm$ 3.9 & 81.3 $\pm$ 3.2 & 79.6 $\pm$ 0.8 & 82.7 $\pm$ 0.7 &  \underline{80.4 $\pm$ 1.2} & 80.7 $\pm$ 1.1 & 82.1 $\pm$ 0.4 & 82.4 $\pm$ 0.4 & 81.6 $\pm$ 0.8 & 81.9 $\pm$ 0.7 \\
    \midrule
    Ours  & C$^2$EFND & \textbf{84.4 $\pm$ 1.2}  & \textbf{89.7 $\pm$ 1.5}  & \textbf{84.0 $\pm$ 1.5}  & \textbf{89.0 $\pm$ 1.2}  & \textbf{83.5 $\pm$ 1.3}  & \textbf{88.5 $\pm$ 1.6}  & \textbf{83.0 $\pm$ 0.9}  & \textbf{ 88.0 $\pm$ 0.7} &  \underline{82.5 $\pm$ 1.1}& \textbf{87.0 $\pm$ 1.3}  \\
    \bottomrule
    \end{tabular}%
    }
  \label{pheme_all}%
    \vspace{-4mm}
\end{table*}%

\begin{table*}[t]
  \centering
  \caption{Performance of methods on \textbf{all} events on \textbf{Twitter16}. Best results are in \textbf{bold} and second best results are \underline{underlined}.}
    \vspace{-2mm}
  \resizebox{1.0\linewidth}{!}{
    \begin{tabular}{c|l|rr|rr|rr|rr|rr}
    \toprule
    \multicolumn{1}{c|}{\multirow{2}[4]{*}{Category}} & \multicolumn{1}{c|}{\multirow{2}[4]{*}{Method}} & \multicolumn{2}{c|}{E1} & \multicolumn{2}{c|}{E2} & \multicolumn{2}{c|}{E3} & \multicolumn{2}{c|}{E4} & \multicolumn{2}{c}{E5} \\
\cmidrule{3-12}          &       & \multicolumn{1}{c}{Accuracy} & \multicolumn{1}{c|}{F1-Score} & \multicolumn{1}{c}{Accuracy} & \multicolumn{1}{c|}{F1-Score} & \multicolumn{1}{c}{Accuracy} & \multicolumn{1}{c|}{F1-Score} & \multicolumn{1}{c}{Accuracy} & \multicolumn{1}{c|}{F1-Score} & \multicolumn{1}{c}{Accuracy} & \multicolumn{1}{c}{F1-Score} \\
    \midrule
    \multirow{4}[2]{*}{\makecell[c]{Single-Event \\ SLM}} & RoBERTa & 73.5 $\pm$ 0.4     & 69.6 $\pm$ 1.4       & 74.1 $\pm$ 0.6    & 72.5 $\pm$ 2.1     &  73.0 $\pm$ 1.0        & 73.0 $\pm$ 1.2       & 76.3 $\pm$ 0.2      & 79.1 $\pm$ 0.8      & 74.8 $\pm$ 0.5      &73.0 $\pm$ 1.1  \\
          & MVAE  & 76.8  $\pm$  0.5 & 71.2  $\pm$  1.2           & 84.2 $\pm$ 0.7 & 84.1 $\pm$ 1.2 & 85.1 $\pm$ 1.0 & 85.0 $\pm$ 1.4 & 85.8 $\pm$ 0.6 & 85.7 $\pm$ 0.8 & 85.5 $\pm$ 0.7 & 85.3 $\pm$ 1.0 \\
          & CompNet & 78.3  $\pm$  0.6 & 73.4  $\pm$  1.1       & 84.4 $\pm$ 0.8 & 84.2 $\pm$ 1.2 & 85.2 $\pm$ 0.9 & 85.1 $\pm$ 1.3 & 85.9 $\pm$ 0.7 & 85.9 $\pm$ 0.7 & 85.7 $\pm$ 0.6 & 85.5 $\pm$ 1.1 \\
          & FTT   & 79.2  $\pm$  0.6 &  \textbf{88.7  $\pm$  0.8}       & 84.5 $\pm$ 1.0 & 84.3 $\pm$ 1.5 & 85.3 $\pm$ 1.0 & 85.2 $\pm$ 1.8 & 85.9 $\pm$ 0.5 & 85.8 $\pm$ 0.6 & 85.8 $\pm$ 0.6 & 85.6 $\pm$ 1.0 \\
    \midrule
\multicolumn{1}{c|}{\multirow{2}[2]{*}{LLM+SLM}} & ARG   & \underline{85.0 $\pm$ 0.7}    &85.0 $\pm$ 0.7       & 84.7 $\pm$ 1.2      &  84.7 $\pm$ 1.2     & 85.8 $\pm$ 0.5      & 85.8 $\pm$ 0.5      &  86.4 $\pm$ 0.2     &  86.4 $\pm$ 0.2     &  87.4 $\pm$ 0.2     &87.3 $\pm$ 0.2  \\
          & EFND   &83.9 $\pm$ 0.2       &  83.8 $\pm$ 0.2       &      84.7 $\pm$ 0.3   & 84.7 $\pm$ 0.3    &  85.3 $\pm$ 0.1     & 85.3 $\pm$ 0.1      & 86.0 $\pm$ 0.4      &  86.0 $\pm$ 0.4     &   86.3 $\pm$ 0.8    & 86.1 $\pm$ 0.8 \\
    \midrule
    \multirow{4}[2]{*}{\makecell[c]{Multi-Domain \\ SLM}} & EANN   & 82.6 $\pm$ 0.6 & 82.4 $\pm$ 0.4 & 85.2 $\pm$ 0.8 & 86.0 $\pm$ 0.4 & 85.4 $\pm$ 0.4 & 86.8 $\pm$ 0.3 & 85.8 $\pm$ 0.4 & 87.9 $\pm$ 0.1 & 85.9 $\pm$ 0.4 & 85.9 $\pm$ 0.4 \\
& MDFEND & 81.4 $\pm$ 0.4 & 82.6 $\pm$ 0.4 & 85.5 $\pm$ 0.4 & 86.1 $\pm$ 0.3 & 85.6 $\pm$ 0.3 & \underline{86.9 $\pm$ 0.3} & 86.4 $\pm$ 0.4 & \underline{88.1 $\pm$ 0.2} & 86.0 $\pm$ 0.2 & 86.2 $\pm$ 0.1 \\
& M3FEND & 80.4 $\pm$ 1.7 & 80.2 $\pm$ 0.2 & 84.5 $\pm$ 1.3 & 85.3 $\pm$ 0.6 & 83.2 $\pm$ 1.1 & 85.6 $\pm$ 0.8 & 85.8 $\pm$ 0.4 & 87.5 $\pm$ 0.4 & 87.2 $\pm$ 0.3 & 84.6 $\pm$ 0.3 \\
& CANMD  & 84.7 $\pm$ 1.0 & 84.7 $\pm$ 1.0 & \underline{86.6 $\pm$ 0.8} & \underline{86.6 $\pm$ 0.8} & \underline{86.3 $\pm$ 0.9}& 86.4 $\pm$ 0.8 & \underline{87.5 $\pm$ 0.5} & 87.5 $\pm$ 0.5 & \underline{87.6 $\pm$ 0.9} & \underline{87.6 $\pm$ 0.9} \\
    \midrule
    Ours  & C$^2$EFND   &  \textbf{87.7 $\pm$ 1.2}      &\underline{86.7 $\pm$ 1.0}       &  \textbf{88.5 $\pm$ 1.5} &  \textbf{88.0 $\pm$ 1.2} &  \textbf{87.5 $\pm$ 1.3} &  \textbf{88.8 $\pm$ 1.6} &  \textbf{88.0 $\pm$ 0.9} &  \textbf{89.4 $\pm$ 0.7} &  \textbf{88.2 $\pm$ 1.1} &  \textbf{88.5 $\pm$ 1.3} \\
    \bottomrule
    \end{tabular}%
    }
  \label{twitter_all}%
    \vspace{-4mm}
\end{table*}%

\subsubsection{Results on all Events}
Due to the phenomenon of rumor resurgence, fake news detection models should not only be capable of identifying emerging events but also retain the ability to detect past events. To this end, we evaluate the model’s performance across all previously encountered events. As observed, our model successfully preserves the ability to detect past events, which can be attributed to the use of distillation loss $\mathcal{L}_{dil}$ and the effective management of the memory bank for replay loss $\mathcal{L}_{replay}$. It is noteworthy that, for other baseline models, we adopt full replay learning, in which all historical training data are utilized during retraining—a process that is highly time-consuming in practice. In contrast, our method employs a memory bank with limited capacity for replay, yet still achieves superior performance in most cases. This demonstrates the practicality of our approach, as it enables model updates with minimal training cost and time.

\subsection{Ablation Study}
To assess the contribution of each module in C$^2$EFND, ablation experiments are conducted and the results are shown in Table~\ref{ablation_study}. ``w/o $\mathcal{L}_{edit}$'' denotes the implementation without knowledge edit.
The results proves that editing up-to-date knowledge into the LLM is essential, which significantly enhances its reasoning ability.
We further examine the continue learning mechanisms.``w/o $\mathcal{L}_{dil}$'' denotes the model without $\mathcal{L}_{dil}$. Removing $\mathcal{L}_{dil}$ leads to a noticeable drop in detecting earlier events,  as the model begins to forget previously learned patterns.
The effectiveness of components used in LLM's inference and SLM's inference is then examined by experiments.
``w/o ICL'' represents LLM inference without demonstrations. ``w/o Rationale'' and ``w/o Knowledge'' denote the removal of rationale and knowledge components from our system.
Notably, all three components are crucial for our framework C$^2$EFND.
Finally, we evaluate the effect of our two-stage active learning strategy.
``w/o active learning'' represents replacing active learning module with random labeling. The decline in model metrics represents the significance of adopting active learning to obtain more important samples in the case of limited labeling.

\subsection{Knowledge Edit Analysis}
\begin{table}[htbp]
\vspace{-4mm}
  \centering
  \caption{Performance of frozen LLM and Knowledge Edited LLM.}
  \vspace{-2mm}
    \begin{tabular}{l|cc|cc}
    \toprule
    \multicolumn{1}{c|}{\multirow{2}[4]{*}{Setting}} & \multicolumn{2}{c|}{Twitter} & \multicolumn{2}{c}{Pheme} \\
\cmidrule{2-5}          & \multicolumn{1}{l}{Accuracy} & \multicolumn{1}{l|}{F1-Score} & \multicolumn{1}{l}{Accuracy} & \multicolumn{1}{l}{F1-Score} \\
    \midrule
    zero-shot & 0.610      & 0.471      & 0.442      & 0.518 \\
    few-shot & 0.660      &  0.706     & 0.604      & 0.739 \\
    ICL & 0.672      & 0.725      & 0.659      & 0.761 \\
    \midrule
    zero-shot+KE &  0.618     & 0.475      & 0.453      & 0.531 \\
    few-shot+KE & 0.681      & 0.732      & 0.616      & 0.751 \\
    ICL+KE & 0.706      & 0.759      & 0.702      & 0.797 \\
    \bottomrule
    \end{tabular}%
  \label{llm_result}%
  \vspace{-3mm}
\end{table}%

One of the core contributions of C$^2$EFND is the lifelong knowledge editing of the LLM. To quantify its effect, we compare three configurations of frozen LLM and edited LLM: zero-shot prompting, few-shot prompting with randomly select four real news and fake news as examples, and in-context learning with eight retrieved demonstrations. Table~\ref{llm_result} presents the accuracy and F1-score of the LLM under these settings on the overall performance of all events. The knowledge-edited LLM clearly outperforms the other configurations, validating the efficacy of our editing approach.

In the zero-shot settings, LLM performs the worst, barely above random guess in some cases, which aligns with previous studies~\cite{hu2024bad}. Providing few-shot examples yields a notable improvement, which indicates that LLMs can partially learn the task criteria from examples.  In-context learning further enhances the LLM’s judgment ability by supplying it with retrieved similar contexts related to each news item. The improvement suggests that when the LLM is provided the right pieces of event knowledge, it can reason more effectively about the veracity of claims. Nonetheless, in-context strategy still has limitations: the LLM is drawing on its fixed weights to interpret the prompt, remains highly reliant on the given demonstrations.

In contrast, our knowledge-edited LLM integrates the labeled news samples and knowledge directly into its model weights before performing the classification task. As illustrated in Table~\ref{llm_result}, the knowledge-edited LLM consistently outperforms its unedited counterparts in all three scenarios. This demonstrates that knowledge editing, by modifying only a small subset of model parameters, enables the LLM to achieve a more fundamental understanding of news events, thereby producing more reasonable inference outcomes.

\begin{table*}[t]
  \centering
  \caption{Ablation Study of C$^2$EFND on all events of Pheme.}
  \vspace{-2mm}
  \resizebox{1.0\linewidth}{!}{
    \begin{tabular}{l|rr|rr|rr|rr|rr}
    \toprule
    \multicolumn{1}{c|}{\multirow{2}[4]{*}{Variant Models}} & \multicolumn{2}{c|}{E1} & \multicolumn{2}{c|}{E2} & \multicolumn{2}{c|}{E3} & \multicolumn{2}{c|}{E4} & \multicolumn{2}{c}{E5} \\
\cmidrule{2-11}          & \multicolumn{1}{c}{Accuracy} & \multicolumn{1}{c|}{F1-Score} & \multicolumn{1}{c}{Accuracy} & \multicolumn{1}{c|}{F1-Score} & \multicolumn{1}{c}{Accuracy} & \multicolumn{1}{c|}{F1-Score} & \multicolumn{1}{c}{Accuracy} & \multicolumn{1}{c|}{F1-Score} & \multicolumn{1}{c}{Accuracy} & \multicolumn{1}{c}{F1-Score} \\
    \midrule
    \multicolumn{1}{c|}{C$^2$EFND}  & 84.4  $\pm$  1.2 & 89.7  $\pm$  1.5 & 84.0  $\pm$  1.5 & 89.0  $\pm$  1.2 & 83.5  $\pm$  1.3 & 88.5  $\pm$  1.6 & 83.0  $\pm$  0.9 & 88.0  $\pm$  0.7 & 82.5  $\pm$  1.1 & 87.0  $\pm$  1.3  \\
    \midrule
w/o $\mathcal{L}_{edit}$   & 82.0 $\pm$ 1.0 & 87.5 $\pm$ 1.3 & 81.5 $\pm$ 1.3 & 88.0 $\pm$ 1.1 & 80.8 $\pm$ 1.2 & 87.0 $\pm$ 1.4 & 80.3 $\pm$ 0.9 & 86.5 $\pm$ 1.0 & 79.9 $\pm$ 1.2 & 85.5 $\pm$ 1.4 \\
w/o $\mathcal{L}_{dil}$    & 83.5 $\pm$ 1.0 & 89.0 $\pm$ 1.4 & 83.0 $\pm$ 1.3 & 88.2 $\pm$ 1.0 & 82.5 $\pm$ 1.2 & 87.8 $\pm$ 1.3 & 82.2 $\pm$ 0.9 & 87.5 $\pm$ 0.7 & 81.8 $\pm$ 1.0 & 86.2 $\pm$ 1.2 \\
\midrule
w/o ICL     & 80.0 $\pm$ 1.1 & 85.0 $\pm$ 1.4 & 78.8 $\pm$ 1.3 & 84.0 $\pm$ 1.1 & 79.2 $\pm$ 1.2 & 83.0 $\pm$ 1.5 & 78.8 $\pm$ 0.9 & 82.5 $\pm$ 0.7 & 78.5 $\pm$ 1.1 & 82.0 $\pm$ 1.3 \\
w/o Rationale                & 83.0 $\pm$ 1.1 & 88.0 $\pm$ 1.5 & 82.0 $\pm$ 1.4 & 87.0 $\pm$ 1.2 & 81.5 $\pm$ 1.3 & 86.5 $\pm$ 1.6 & 81.0 $\pm$ 1.0 & 86.0 $\pm$ 0.7 & 80.5 $\pm$ 1.2 & 85.0 $\pm$ 1.3 \\
w/o Knowledge                & 82.5 $\pm$ 1.2 & 87.5 $\pm$ 1.5 & 81.8 $\pm$ 1.5 & 86.8 $\pm$ 1.2 & 81.0 $\pm$ 1.4 & 86.0 $\pm$ 1.6 & 80.5 $\pm$ 0.9 & 85.5 $\pm$ 0.7 & 80.0 $\pm$ 1.2 & 85.0 $\pm$ 1.3 \\
\midrule
w/o active learning          & 82.8 $\pm$ 1.0 & 88.8 $\pm$ 1.3 & 82.0 $\pm$ 1.3 & 87.5 $\pm$ 1.0 & 81.5 $\pm$ 1.2 & 86.8 $\pm$ 1.4 & 81.0 $\pm$ 0.8 & 86.2 $\pm$ 0.6 & 80.8 $\pm$ 1.1 & 85.8 $\pm$ 1.2 \\
    \bottomrule
    \end{tabular}%
    }
  \label{ablation_study}%
  \vspace{-4mm}
\end{table*}%
\subsection{Multi-Round Learning Analysis}
\begin{table}[htbp]
  \centering
  \vspace{-2mm}
  \caption{Analysis on multi-round learning on Twitter16 E1.}
  \vspace{-2mm}
    \begin{tabular}{c|c|cc|cc}
    \toprule
    Round & Model & \multicolumn{1}{c}{Accuracy} & \multicolumn{1}{c|}{F1-Score} & \multicolumn{1}{c}{Clean Pool} & \multicolumn{1}{c}{Noisy Pool} \\
    \midrule
    \multirow{2}[2]{*}{1} & SLM   & 80.2      & 79.4      & 4633      & 2795 \\
          & LLM   & 67.5      & 70.2      &  4633     & 2795 \\
    \midrule
    \multirow{2}[2]{*}{2} & SLM   & 84.0      & 82.5      & 5751      & 1677 \\
          & LLM   & 68.1      & 71.4      & 5751      & 1677 \\
    \midrule
    \multirow{2}[2]{*}{3} & SLM   & 88.1      & 87.2      & 6459      & 969 \\
          & LLM   &  69.3     & 72.7      & 6459      & 969 \\
    \midrule
    \multirow{2}[2]{*}{4} & SLM   &  86.7     & 85.9      & 6640      & 788 \\
          & LLM   & 69.6      & 73.1      &  6640     & 788 \\
    \midrule
    \multirow{2}[2]{*}{5} & SLM   & 86.1      & 85.2      & 6692      & 736  \\
          & LLM   & 68.9      & 72.3      & 6692      & 736 \\
    \bottomrule
    \end{tabular}%
  \label{tab:multi_round}%
\end{table}%

To verify the effectiveness of the iterative multi-round learning framework on data filtering and model improvement. We conduct experiments on the inference results of SLMs and LLMs on Twitter E1 and record the number of samples in clean data pool and noisy data pool as shown in Table~\ref{tab:multi_round}. Notably, the first three rounds produce substantial gains: the SLM’s accuracy climbs from about 80.2\% in Round 1 to 88.1\% by Round 3, accompanied by a similar F1-score improvement. This boost corresponds with a sharp increase in the clean data pool by $2^{nd}$ active learning and data filtering. These results confirm that most of the informative unlabeled samples are correctly identified and added in the early rounds, which markedly improves the model’s discrimination ability. By Round 3, the model has leveraged almost all high-confidence data, and its performance on the new event nearly saturates.

Increasing the rounds further yields diminishing returns and can even trigger a slight performance drop. As seen in Table X, an additional fourth round causes a modest decrease in SLM accuracy (from 88.1\% down to 86.7\%). We attribute this to overfitting with noisy samples: once nearly all relevant samples have been included by Round 3, extra rounds are more likely to incorporate marginal or misclassified samples. This can slightly erode the previously gained accuracy. 

\subsection{Sparse Data Analysis}
\begin{table}[htbp]
\vspace{-4mm}
  \centering
  \caption{Sparse data analysis on Twitter16 and Pheme.}
  \vspace{-2mm}
  \resizebox{1.0\linewidth}{!}{
    \begin{tabular}{c|l|rr|rr}
    \toprule
    \multirow{2}[4]{*}{Label Ratio} & \multirow{2}[4]{*}{Method} & \multicolumn{2}{c|}{Twitter} & \multicolumn{2}{c}{Pheme} \\ 
    \cmidrule{3-6}
           &       & Accuracy & F1-Score & Accuracy & F1-Score \\
    \midrule
    \multirow{3}[2]{*}{5\%}
      & ARG      & 75.0 $\pm$ 0.9 & 74.8 $\pm$ 1.2 & 74.0 $\pm$ 1.1 & 72.5 $\pm$ 1.3 \\
      & M3FEND   & 75.8 $\pm$ 0.8 & 75.5 $\pm$ 1.0 & 74.7 $\pm$ 1.2 & 73.0 $\pm$ 1.4 \\
      & C$^2$EFND & 81.5 $\pm$ 0.7 & 81.1 $\pm$ 0.9 & 79.2 $\pm$ 0.9 & 77.8 $\pm$ 1.0 \\
    \midrule
    \multirow{3}[2]{*}{10\%}
      & ARG      & 81.0 $\pm$ 0.7 & 80.8 $\pm$ 1.0 & 77.6 $\pm$ 1.1 & 75.5 $\pm$ 1.3 \\
      & M3FEND   & 81.3 $\pm$ 0.8 & 80.0 $\pm$ 1.1 & 78.1 $\pm$ 1.0 & 76.2 $\pm$ 1.2 \\
      & C$^2$EFND & 83.8 $\pm$ 0.7 & 83.2 $\pm$ 0.9 & 80.1 $\pm$ 1.0 & 78.5 $\pm$ 1.1 \\
    \midrule
    \multirow{3}[2]{*}{15\%}
      & ARG      & 84.0 $\pm$ 0.6 & 83.9 $\pm$ 0.8 & 80.9 $\pm$ 1.0 & 79.2 $\pm$ 1.3 \\
      & M3FEND   & 83.4 $\pm$ 0.7 & 81.2 $\pm$ 0.9 & 80.3 $\pm$ 1.1 & 78.5 $\pm$ 1.2 \\
      & C$^2$EFND & 86.5 $\pm$ 0.5 & 86.0 $\pm$ 0.7 & 81.8 $\pm$ 0.8 & 80.9 $\pm$ 1.0 \\
    \midrule
    \multirow{3}[2]{*}{20\%}
      & ARG      & 87.4 $\pm$ 0.2 & 87.3 $\pm$ 0.2 & 83.2 $\pm$ 0.6 & 81.4 $\pm$ 0.6 \\
      & M3FEND   & 87.2 $\pm$ 0.3 & 84.6 $\pm$ 0.3 & 83.1 $\pm$ 0.8 & 81.2 $\pm$ 1.0 \\
      & C$^2$EFND & 88.2 $\pm$ 1.1 & 88.5 $\pm$ 1.3 & 82.5 $\pm$ 1.1 & 87.0 $\pm$ 1.3 \\
    \bottomrule
    \end{tabular}%
  }
  \label{tab:sparse_analysis}%
\end{table}

To evaluate robustness under limited annotation, we compare C$^2$EFND agains ARG~\cite{hu2024bad} and M3FEND~\cite{zhu2022memory} on both Twitter and Pheme datasets as illustrated in Table~\ref {tab:sparse_analysis}. We select 5\%, 10\%, and 15\% of the data from the  Twitter and Pheme datasets as annotated samples. The accuracy and f1-score on all events are used as the metrics.
Notably, with more limited data labeled, C$^2$EFND achieves significantly higher accuracy and F1-score than all competitors. For instance, under this low-label scenario(e.g. 5\% labeled data), our model’s F1 remains high, whereas M3FEND and ARG suffer substantial performance degradation. This highlights that C2EFND can effectively leverage unlabeled data and pre-trained knowledge to compensate for labeling scarcity. The advantage stems from our two-stage active sampling and multi-round self-training, which ensure the model quickly grasps core event features even from a handful of labeled examples.

\subsection{Parameter Sensitivity Analysis}
\begin{figure}[h]
\centering
\includegraphics[scale=0.9]{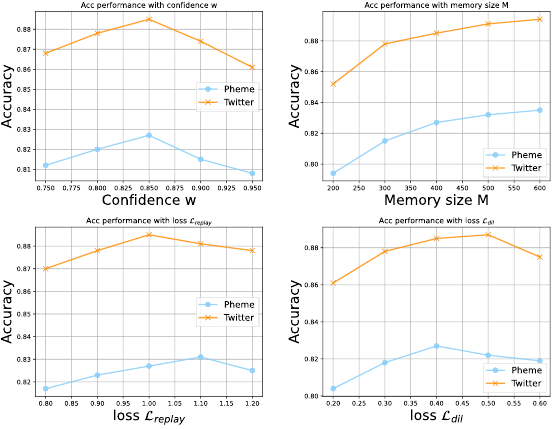}
    \setlength{\abovecaptionskip}{-2mm}
    \setlength{\belowcaptionskip}{-5mm}
    \caption{Hyper-parameter sensitivity analysis on Twitter16 and Pheme.}
    \label{fig:hyperparameter}
\end{figure}

We finally investigate the model's sensitivity to several key hyperparameters: the confidence threshold $\omega$ for pseudo-labeling, the replay memory size $M_{max}$, and the loss weights $\lambda_{1}$ and $\lambda_{2}$ in the training objective on all events of Twitter16 and Pheme. The results are shown in Figure~\ref{fig:hyperparameter}.

For the confidence threshold $\omega$: a lower $\omega$ makes the criteria too lenient, allowing many noisy or incorrect pseudo-labels into the training pool, resulting in decrease on model performance. On the other hand, a very high $\omega$ is overly conservative, too few pseudo-labeled instances are accepted as clean, which limits the amount of training data for the SLMs to learn enough events knowledge. For the memory size $M_{max}$: a too small memory size hurts performance on past events, signifying some degree of catastrophic forgetting as expected. However, we observe diminishing returns with very large memory sizes. Beyond a certain point, increasing the memory capacity (toward using all past samples) yields little to no performance gain, yet greatly increases computational cost. This confirms that a strategically maintained, moderate-sized replay buffer is sufficient for continual fake news detection.

For the loss weight $\lambda_1$ and $\lambda_2$, these two losses are used to balance learning between past events and emergent events. When the values of these two loss terms are too small, the model's performance on past events deteriorates significantly. This is primarily due to insufficient learning on historical data and weak regularization from the previous model. Conversely, when these loss values are too large, the model tends to overemphasize past events but fails to learn from emergent events, resulting in decline performance in all events.

\section{Conclusion}
In conclusion, we introduce a novel and practical continuous emergent fake news detection setting and propose a novel C$^2$EFND framework to solve this problem.
By addressing the core challenges—limited annotations, continuous knowledge updates, and collaborative inference—C$^2$EFND achieves superior performance on two benchmark datasets.
The proposed multi-round learning efficiently selects pseudo labels for training, and the lifelong knowledge editing module effectively integrates evolving news into LLMs.
Furthermore, the replay-based continual learning approach successfully prevents catastrophic forgetting in SLMs.
Future research will extend the framework’s applicability to more complex multimodal scenarios and further optimize computational efficiency.

\bibliographystyle{ieeetr}
\bibliography{sample_reduce}

\end{document}